\documentclass[10pt,letterpaper, conference]{IEEEtran}
\newcommand{\roms}[1]{\textcolor{blue}{}}
\newcommand{\fred}[1]{\textcolor{red}{}}

\input{LaTeX_macros_bib/alejandro_symbol.sty}

\usepackage{csvsimple}
\usepackage{booktabs}
\usepackage{comment}
\usepackage{array}
\usepackage{stfloats}
\usepackage{upgreek}
\usepackage{LaTeX_macros_bib/saurav_macros}
\usepackage{LaTeX_macros_bib/shorthands_minimal}
\usepackage{siunitx}
\usepackage[caption=false,font=footnotesize]{subfig}
\usepackage{rotating}
\usepackage{cite}
\usepackage{enumitem}

\usepackage{pgfplots}
\pgfplotsset{width=0.49\textwidth,compat=1.18}
\usetikzlibrary{pgfplots.statistics}
\usepackage{adjustbox}
\usepackage{xurl}  %

\newcommand{\ResponseRed}[1]{{}}
\newcommand{\ResponseGreen}[1]{{}}
\newcommand{\ResponseBlue}[1]{{}}

\usepackage{color}
\newcommand{\sa}[1]{\textcolor{mBlue}{}}
\usepackage[normalem]{ulem}
\PassOptionsToPackage{hyphens}{url}
\usepackage{url}

\IEEEoverridecommandlockouts
\author{Saurav Agarwal$^{*}$, Frederic Vatnsdal$^{*}$, Romina Garcia Camargo,\\Carlos Nieto-Granda, Vijay Kumar, and Alejandro Ribeiro%
  \thanks{$^{*}$Equal contribution.\newline \mbox{F.~Vatnsdal, R.~G.~Camargo, V.~Kumar}, and \mbox{A.~Ribeiro} are with the University of Pennsylvania, USA.}%
\thanks{Email: {\tt\{vatnsdal,rominag,kumar,aribeiro\}@\newline seas.upenn.edu}.\newline
\mbox{S.~Agarwal} is with the Dept. of Mechanical Engineering, Indian Institute of Technology Bombay, India. {Email: \tt saurav.agarwal@iitb.ac.in}.\newline C.~Nieto-Granda is with the U.S. Army DEVCOM Army Research Laboratory, Adelphi, MD, USA. {Email: \tt carlos.p.nieto2.civ@army.mil}.}}

\begin{document}
\bstctlcite{IEEEexample:BSTcontrol}
\title{\vspace{0.25in}\Large \bf A Scalable Multi-Robot Framework for Decentralized and Asynchronous Perception-Action-Communication Loops}
\maketitle
\thispagestyle{plain}
\pagestyle{plain}
\begin{abstract}
We develop a decentralized Perception-Action-Communication (PAC) system for multi-robot teams that enables them to collaborate in large scale, outdoor environments. Our system natively supports deployments at any scale by leveraging a graph neural network (GNN) to diffuse information hop-by-hop across the fleet's network. This achieves global collaboration from individual robots limited to local sensing and communication. Fully asynchronous, the core modules of PAC: perception, inter-robot communication, message aggregation and action are clocked at different frequencies with information flowing between them through buffers. We implement the PAC system as a series of highly extensible ROS2 nodes to serve as the foundational infrastructure for deployable swarm systems. PAC is validated in the real world with outdoor experiments with up to $N=20$ quadrotor robots and in simulations based on real-world data with up to $N=100$. These validations show that our system upholds crucial properties for field-deployable robot collectives: scalability, resiliency and repeatability.

\end{abstract}

\section{Introduction}
\label{sc:intro}

Perception-Action-Communication (PAC) loops facilitate decentralized multiagent collaboration toward shared global objectives \cite{agarwal_2024, hu2022}.
By processing local perception inputs and incoming messages to synthesize control commands and outbound messages, individual agents create a \emph{global} control loop grounded in \emph{local} communication.
Graph Neural Networks (GNNs) \cite{KipfW17, bruna2014spectral, deferrard2016convolutionalnnongraphs} are an established tool for learning such multi-robot policies, as inter-agent interaction topologies naturally map to graph structures  (Section \ref{sc:gnn}).
From early demonstrations in flocking~\cite{TolstayaPMLR20, GamaDecentralized2022, chen2024learning}, GNN-driven decentralized control has expanded to complex applications such as coverage control~\cite{Gosrich22} and target tracking~\cite{zhou2022, rangesh2021trackmpnn}.
This success is underpinned by theoretical stability and transferability properties, explaining their capacity to generalize across dynamic network topologies and team size \cite{RuizGR21}.

\begin{figure}[tbp]
	\centering
    \vspace{-12pt}
	\includegraphics[width=1.0\linewidth]{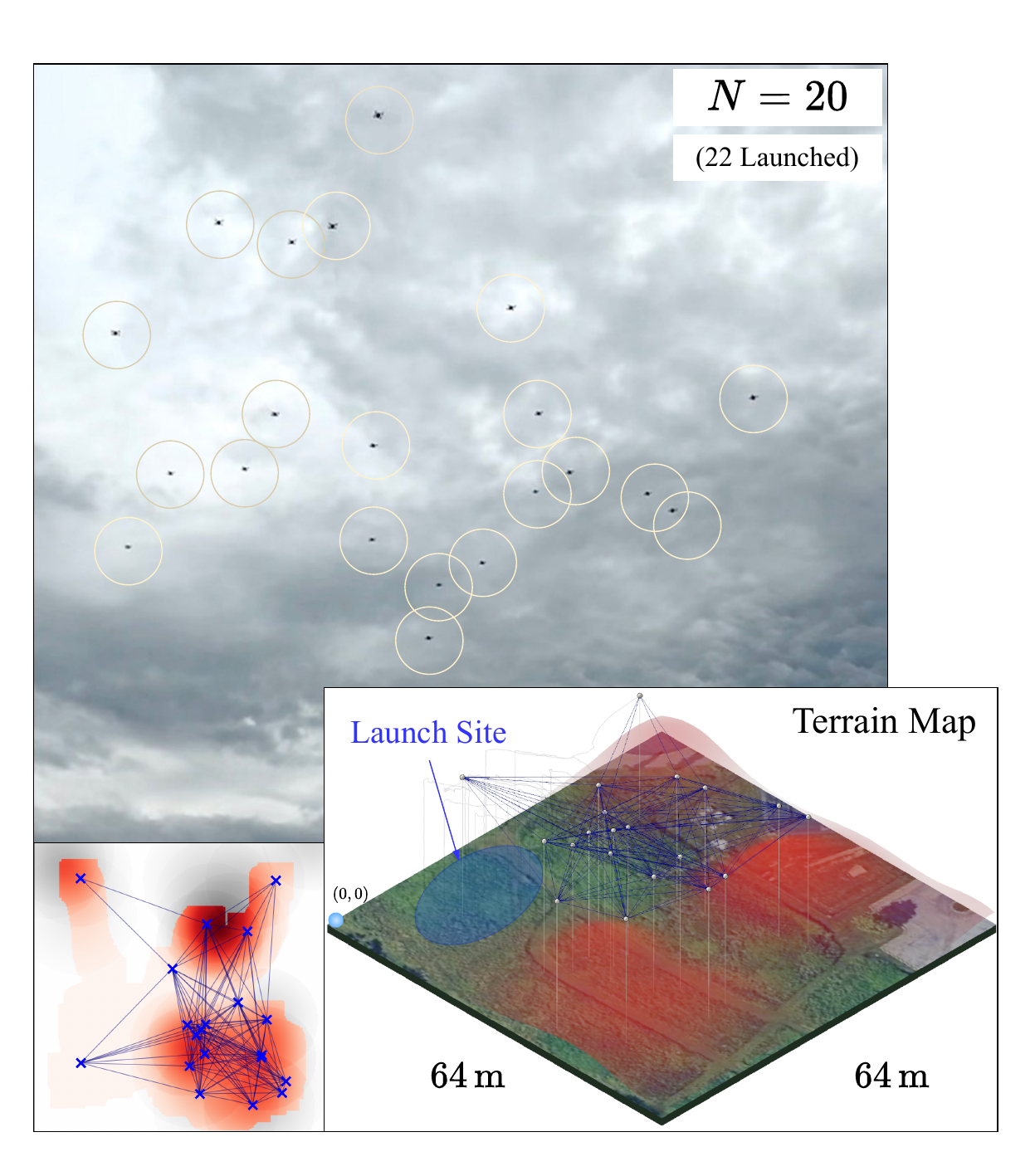}
	\caption{Perception-Action-Communication (PAC):
        \textbf{Top:} We validate the proposed PAC system with real-world experiments on $N=20$ quadrotor platforms with size, weight, and power (SWaP)-limited sensing, computation, and communication capabilities.
        \textbf{Bottom Left:} PAC enables decentralized collaborative control of robots (blue), driving them to autonomously uncover hidden priority fields (red) and minimize the cost of covering them.
        \textbf{Bottom Right:} Robots are launched from the ground in a tight cluster before fanning out over the area once they reach operational altitude.}
	\label{fig:moneyshot}
\end{figure}

\begin{figure*}[tbp]
	\centering
	\includegraphics[width=1.0\linewidth]{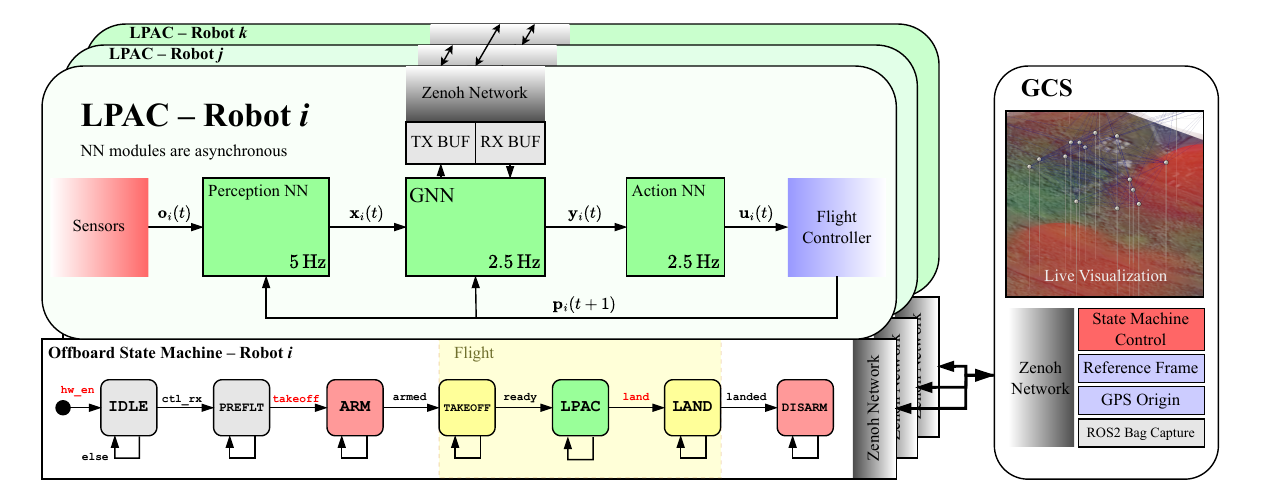}
	\caption{%
        \textbf{Top left:} Perception-Action-Communication (PAC) loops using an architecture composed of three Neural Networks (NN):
		The perception NN, usually a Convolution NN, computes features using the sensor data obtained through perception.
		The core of the architecture is a collaborative NN, a Graph NN in our case, which computes the messages to be exchanged between robots and aggregates received messages.
	    The GNN also computes features for the action NN, often a shallow Multi-Layer Perceptron (MLP), the output of which is sent to a control module such as a low-level flight controller.
        \textbf{Bottom left:} each robot runs an offboard control state machine that manages both preparatory maneuvers and PAC execution during trials.
        \textbf{Right:} the ground control station (GCS) enables us to manage the state machines of the entire fleet remotely for coordinated state transitions.
        The GCS serves as a control panel with emergency handling, live visualizer of the fleet's state, reference frame broadcasting, and data collection.
        Internally, software functionality is packaged as ROS2 nodes that communicate across the fleet and with the GCS through the Zenoh network protocol.}
	\label{fig:overview}
\end{figure*}

While existing frameworks~\cite{BlumenkampGNN2022,Tzes_icra_2023} consider GNNs for decentralized multi-robot collaboration, they do not provide large-scale field deployments with a fully asynchronous framework.
This is the main contribution of our work:

\begin{itemize}[leftmargin=2.3em, labelsep=0.3em, itemindent=0pt]

\item[\textbf{(C1)}] We develop the first implementation of a \textbf{large-scale learned PAC system} that implements policies learned by a GNN (illustrated in Figures~\ref{fig:moneyshot} and~\ref{fig:overview}).

\end{itemize}

\noindent Through the implementation of four asynchronous modules to handle perception, inter-robot communication, message aggregation and filtering, and low-level control, we enable concurrent execution at variable frequencies (Section \ref{sc:framework}).
We provide a ROS2~\cite{ros2}-compatible open-source implementation of the framework, which is deployed independently on each robot as well as in a Ground Control Station (GCS).

We evaluate the performance of the proposed PAC framework on the decentralized coverage control problem~\cite{Cortes05}, where each agent perceives a \emph{local} section of an importance function.
The team must navigate to provide sensory coverage weighted by the importance function---mathematically formulated as minimization of coverage cost.
Through extensive real-world deployments and software in-the-loop simulations we demonstrate the following features (Section~\ref{sc:experiments}):

\begin{itemize}[leftmargin=2.3em, labelsep=0.3em, itemindent=0pt]
  \item[\textbf{(C2)}] \textbf{Global execution from local information}: The performance of the proposed decentralized PAC system closely matches that of the (clairvoyant) centralized system with access to global information (Section~\ref{sec:c1}).
  \item[\textbf{(C3)}] \textbf{Scalability}: The system  maintains performance when tested in out-of-distribution team sizes; up to 20 robots in the real-world and 100 in simulation (Section \ref{sec:c2}).
  \item[\textbf{(C4)}] \textbf{Resilience} to team reconfiguration: The system policy adjusts smoothly when robot teams get reinforced or individual robots get decommissioned (Section \ref{sec:c3}).
  \item[\textbf{(C5)}] \textbf{Robustness} to degraded communication (Section \ref{sec:c4}).
  \item[\textbf{(C6)}] \textbf{Repeatability}: Consistent performance over multiple runs on the same scenario (Section \ref{sec:c5}).
\end{itemize}

\section{Decentralized Graph Neural Networks}
\label{sc:gnn}

We learn distributed collaborative policies with graph neural networks (GNNs)~\cite{KipfW17}, which are layered information processing architectures based on recursive exchanges of messages across neighboring agents (\fgref{fig:gnn}).
Formally, let $\bbS$ be the adjacency matrix that characterizes the connectivity of the team where we allow for weighted entries $S_{ij}$. The GNN's output at Layer $\ell-1$ and Agent $i$ is denoted as $\bbx_i^{\ell-1}$ and taken as input at Layer $\ell$. This output becomes a message $\bbm_{0i}^\ell = \bbx_i^{\ell-1}$ that is exchanged and aggregated with neighbors' messages to produce a sequence of messages $\bbm_{ki}^\ell$ defined by the recursion
\begin{align} \label{eqn_gnn_message_exchange}
    \bbm_{ki}^\ell = \sum_{j \in \ccalN(i)} S_{ij} \bbm_{k-1,j}^\ell .
\end{align}
In this expression the messages $\bbm_{k-1,j}^\ell$ of neighbors $j \in \ccalN(i)$ are inputs to a weighted sum with coefficients determined by the graph weights $S_{ij}$. Because the sum is restricted to neighboring nodes, the computation of the message $\bbm_{ki}^\ell$ is localized and distributed. To compute this message, Agent $i$ requires access to messages $\bbm_{k-1,j}^\ell$ that it can receive from neighbors over the communication graph.

After the sequence of $K+1$ messages $\bbm_{ki}^\ell$ with $k=0,\ldots, K$ is computed, they are scaled by processing matrices $\bbH_k^\ell$, summed and passed through a pointwise nonlinear operation to compute the output of Layer $\ell$,
\begin{align} \label{eqn_gnn_layer_ell_output}
    \bbx_i^\ell = \sigma \bigg[\, \sum_{k=0}^K \bbH_k^\ell \bbm_{ki}^\ell \,\bigg].
\end{align}
If the features of layer $\ell$ have dimension $p_\ell$, the matrices $\bbH_k^\ell$ have $p_\ell \times p_{\ell-1}$ entries. The matrices $\bbH_k^\ell$ for all $k$ and all $\ell$ are the parameters that we train to coordinate the team. 

\begin{figure*}[ht]
\centering
\input{figures/graph_shift_register.tex}
\caption{A Graph Neural Network Layer.
Layer $\ell$ processes the output of layer $\ell-1$ by diffusing information over the team's graph $\bbS$.
Diffused messages $\bbm_{ki}^\ell$ of order $k$ aggregate the messages $\bbm_{k-1,j}^\ell$ of order $k-1$ of neighboring agents [cf. \eqref{eqn_gnn_message_exchange}].
This recursive construction results in a sequence of messages  $\bbm_{k,i}^\ell$ with $k=0,\ldots,K$ that aggregate information from $k$ hops away.
The output of layer $\ell$ is obtained by applying trainable linear maps $\bbH_k^\ell$ to each diffused message and passing their addition through a pointwise nonlinear function [cf. \eqref{eqn_gnn_layer_ell_output}]. \label{fig:gnn}}
\end{figure*}

To complete the definition of a GNN, the input to Layer~1 is $\bbx_i^0 = \bbx_i$ for some given $\bbx_i$ and the output follows after $L$ layer recursions.
For brevity, we let $\bbX = [\bbx_i]_{i}$ be a variable that aggregates the input states of all agents and we further let $\bbH = [\bbH_{k}^{\ell}]_{k\ell}$ aggregate the trainable parameters of all layers. We then denote the output of the GNN as the function map 
\begin{align} \label{eqn_gnn_abstract}
    \bbPhi(\bbX, \bbS; \bbH) = [\bbx_i^{L}]_{i},
\end{align}
that aggregates the outputs $\bbx_i^{L}$ of Layer $L$ across all agents~$i$.

\medskip \noindent \textbf{Imitation of Centralized Policies.} To train GNNs for distributed collaboration, we consider training sets $\ccalT$ made up of triples $(\bbX, \bbS, \bbU)$ where $\bbU$ is a benchmark---ideally optimal---\emph{centralized} control action associated to graph $\bbS$ and system state $\bbX$.
Furthermore, the benchmark can be \emph{clairvoyant} with full observability.
We then search for parameters $\bbH^\star$ that maximize the fit between benchmark actions $\bbU$ and actions $\bbPhi(\bbX, \bbS; \bbH)$ produced by the GNN as evaluated by a given cost function $c(\cdot,\cdot)$ ~\cite{GamaDecentralized2022},
\begin{align} \label{eqn_gnn_training}
    \bbH^\star ~=~ \argmin_\bbH \, 
                   \sum_{(\bbX, \bbS, \bbU)\in \ccalT} 
                       c\Big[\, \bbPhi(\bbX, \bbS; \bbH),\, \bbU \Big] .
\end{align}
It is important that in \eqref{eqn_gnn_training} we train a \emph{decentralized} GNN to imitate a \emph{centralized} benchmark policy. This is possible because reference actions $\bbU$ are available during \emph{offline} training. GNNs excel at discovering decentralized policies that behave similarly to centralized experts.
During decentralized deployment, these policies continue to exhibit global decision making despite operating only on the information available to individual robots, e.g., \cite{agarwal_2024, hu2022, TolstayaPMLR20, GamaDecentralized2022, chen2024learning, Gosrich22, zhou2022, rangesh2021trackmpnn}.

\section{Asynchronous PAC Framework}
\label{sc:framework}

The asynchronous design of the PAC loop enables concurrent execution across modules.
Lightweight modules with lower computational costs can be updated at higher frequencies, reducing overall execution overhead on multi-core processors and thereby enhancing real-time performance.
This is in contrast to prior work, where the perception and GNN computations were considered to be executed synchronously, even if communication and low-level controller are asynchronous, affecting the computation time and the number of message aggregations that can be performed by the GNN module.

The system is physically realized across two domains: the robot (\scref{sc:phys_robot} and \scref{sc:robot-hw}) and the Ground Control Station (\scref{sc:phys_gcs}); \fgref{fig:overview}.
These two domains are staffed by an ensemble of ROS2 nodes run inside Docker containers deployed on the robots and the ground control station, while the Zenoh-based communication network \cite{zenoh2023} serves as the incorporeal substrate.
The use of containers is crucial to the software maintenance and the bring-up of a fleet of robots.

\subsection{Policy}\label{sc:phys_robot}
The framework comprises multiple neural networks in tandem, referred to as a learnable PAC (LPAC) policy~\cite{agarwal_2024}.
This policy is trained with an observation-action dataset generated from a clairvoyant centralized algorithm in simulations.
Instantiated across asynchronous modules, LPAC grants each robot autonomy and serves as the pilot, ingesting sensory readings and positions as the robot traverses the environment, as well as the messages $\bbm_{j}$ and positions $\bfp_j$ received from neighboring robots ($j \in \calN(i)$).
From this input, LPAC produces the velocity $\bfu_i$ that is delivered to the offboard control interface to be ultimately sent to the platform specific low-level controller.

We set the Perception module to refresh the embedding $\bbx_{i}$ at a rate of $\SI{5}{\hertz}$.
In parallel, the GNN-based Communication module queries the message buffer manager, which holds positions and embeddings received from other robots $[\bfm_j, \bfp_j]$, at a rate of $\SI{2.5}{\hertz}$.
Operating the Communications module at a lower frequency is a design decision that allows a greater window for aggregation.
The robot's velocity $\bbu_i$ is read from the output buffer of the Action module and sent to the control interface at a rate of $\SI{10}{\hertz}$.

\medskip \noindent The \textbf{Perception} module processes sensor data $\bbo_i$ into input features for the GNN.
For coverage control, we use a Convolutional Neural Network (CNN) that produces a low dimensional feature vector from local observations of the importance density function.

\medskip \noindent \textbf{Inter-Robot Communication.}
Robots {\em broadcast} their message $\mv m_i$, for robot $i$, to other robots within their communication range $r_c$, i.e., the neighboring robots $\mc N(i)$.
They also receive messages from their neighbors: $\mv m_j\forall j \in \mc N(i)$.
Generally, communication hardware may allow either receiving a message or transmitting a message at a given time.
Thus, the communication module may need to be executed at twice the frequency of the GNN message aggregation module.
We use two buffers to maintain messages: a transmitter buffer T\textsubscript{x}, which stores the message to be transmitted, and a receiver buffer R\textsubscript{x}, which stores the most recent message received from each neighbor.
A message buffer interfaces the GNN to the radio, collecting the most recently received messages from neighbors while also loading the outgoing message for transmission.
Messages are cleared after aggregation to avoid stale messages being consumed in the future.

\medskip \noindent The \textbf{GNN} is the collaborative nexus of the architecture and aggregates incoming messages from neighbors while simultaneously generating messages to be transmitted back to said neighbors.
Because the GNN relies on local communication only, it is both decentralizable and highly scalable and transferable to unseen communication graph topologies.
The GNN enables information to percolate over multiple hops across the network, thereby granting the fleet a larger scope than any single robot can sense individually.

\medskip \noindent The \textbf{Action} module receives the local embeddings inferred by the GNN module ($\bby_i$ in \fgref{fig:overview}) to decide the next action $\bbu_i$ towards the global goal. It is implemented with a shallow Multi-Layer Perceptron (MLP), the output of which is sent to a control module such as a low-level flight controller.
\subsection{Control Interface}\label{sc:control_interface}
The LPAC policy outputs planar velocity actions for the robots.
Altitude control is handled by a separate proportional controller, and vertical velocities are fused into the commanded planar velocity at $\qty{10}{\hertz}$.
The offboard node maintains the operational state-machine of the robot throughout an experiment and provides control input when not executing LPAC's velocity commands.
The state machine receives messages from the GCS mission control node as the control input to progress the robot through a deployment.
Velocities are computed in the mission frame $\calF_M$ and transformed into the vehicle's local mission frame $\calF_L$ and positions are converted in the other direction.
The conversion is computed using a fixed transform from the vehicle's GPS fix at boot and the received mission origin from the GCS; the transform is constructed using a geodesic inverse~\cite{Karney_2012}.
The low-level PX4 interface is exposed to ROS2 through a MicroDDS pipeline~\cite{eProsima_Micro_XRCE_DDS}, used for reading the positions in the vehicle's local frame and sending velocity commands to the robot.

\subsection{Ground Control Station}\label{sc:phys_gcs}
The GCS runs the mission control and visualization nodes that enable the operator to interface with the fleet during experiments.
The GCS distributes the frame of reference and the origin position to the fleet, enabling the team to compute their own relative positions.
Safety measures such as geofencing and mission-pause are available through the GCS.
The GCS has no influence on the behavior of the LPAC policy and is used for safety, experiment setup, and recording.

\section{Autonomous Aerial Vehicle Hardware}
\label{sc:robot-hw}
\begin{figure}[htb]
    \includegraphics[width=\linewidth]{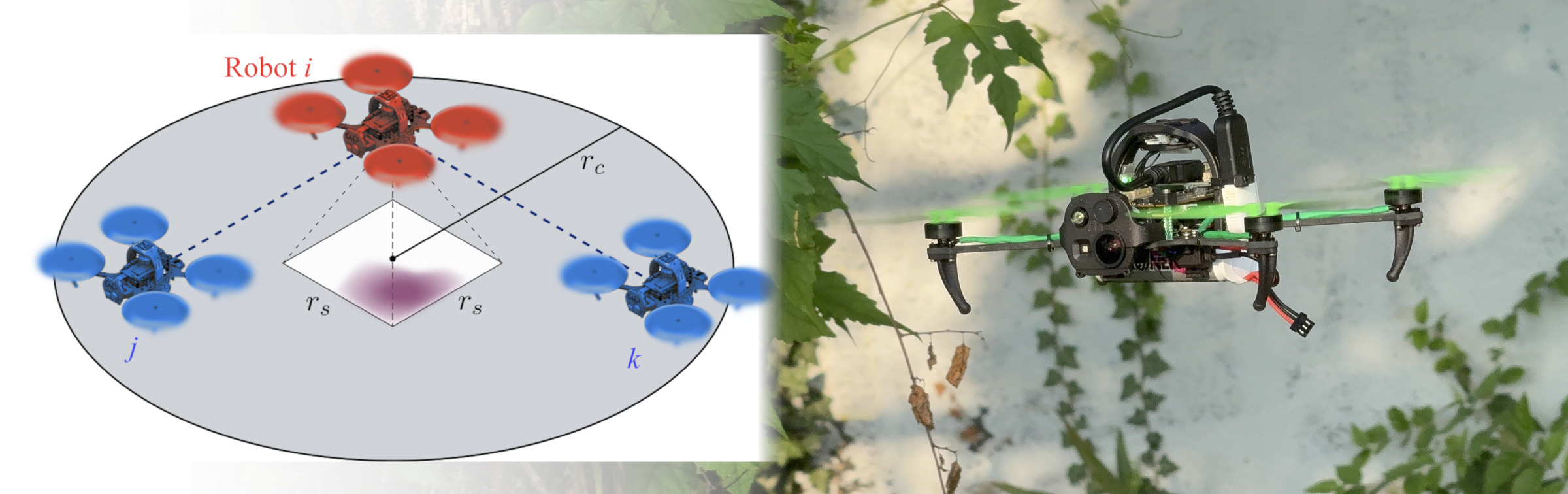}
    \caption{We use the ModalAI Starling 2 as our robot platform. The vehicle has onboard compute running both the flight controller and the PAC system. Localization is accomplished using GNSS as well as inertial and barometric measurements. Inter-vehicle communication is handled by the onboard WiFi radios. \label{fig:starling_hardware}}
\end{figure}
\subsection{Airframe and Compute}
A fleet of up to $20$ of ModalAI's Starling 2 quadrotors serve as our robotics platform for demonstrating the real world efficacy and robustness of PAC framework (\fgref{fig:starling_hardware}).
Our chosen platform is size weight and power (SWaP) limited; the \SI{220}{\milli\meter} airframe weighs only \SI{280}{\gram}~\cite{ModalAI_VOXL2_product}.
From the factory, each Starling 2 is equipped with a Qualcomm QRB5165 processor mounted on a VOXL2 single board computer (SBC).
The 64 bit ARM processor has 8 cores that are able to reach clock speeds of 2.84 GHz and is supported by 8GB of LPDDR5 memory.
LPAC inference and PX4 v1.14 autopilot run in tandem on the same on-board VOXL2 computer.
The VOXL2 runs a modified version of Ubuntu 18.04 with the 4.19.125 kernel.

\subsection{Sensors}
The focus of our experiments is on LPAC as a decentralized control policy and the interaction between agents via embeddings shared over the network.
We operate under the assumption that a computer vision module exists to convert raw sensory observations to importance values using object detection.
Hence, we simulate sensing on-board each robot independently.
Each robot must be able to localize itself in our mission frame $\calF_M$ and maintain a fixed altitude.
The vehicle is equipped with the uBlox-M10S GNSS receiver, QMC5883L compass, two TDK ICM-42688-P inertial measurement units (IMU) and two barometers: a Bosch BMP388 and a TDK ICP-10100.
The PX4 flight controller runs an extended Kalman filter (EKF) for sensor fusion and we configure the EKF to fuse GNSS, IMU and barometer measurements for localization.
The Starling 2 includes several cameras for low-altitude perception, which we disable for our high-altitude experiments; future work will involve incorporating real-time vision into the perception module.

\subsection{Communication}
To facilitate collaboration through communication between agents, each robot is equipped with an ALFA AWUS036EACS WiFi radio \cite{Alfa_AWUS036EACS_product} operating in the \SI{5}{\giga\hertz} band.
The radio has a typical power output of 17 dBm and has an integrated antenna.
Two TP-Link AX1800 WiFi routers manage traffic from vehicle to vehicle and between each vehicle and the GCS, configured in different channels to avoid overcrowding.
All radios in the system support the 802.11ac WiFi standard, enabling a maximum data-rate of 433 Mbps.

\section{Experiments}
\label{sc:experiments}

Coverage control requires a team of $N$ robots to provide sensor coverage to features of interest described by an \emph{importance density function} (IDF)~\cite{Du99, Cortes05,Gosrich22}. Through {\em Voronoi partitions} $\mc P_i$, each robot is assigned to a distinct subregion of the environment.
The objective of the coverage problem is to minimize the cost to cover the environment, $\mc{J}$, weighted by the IDF $\phi(\cdot)$:
\begin{equation} \label{eq:coverage_cost}
    \mc{J}(\mv{p}_{1},\ldots,\mv{p}_{N}) = \sum_{i=1}^{N} \int_{\mc{P}_{i}} \|\mv{p}_{i} - \mv{q}\|^2\phi(\mv{q}) \mv{dq}.
\end{equation} 
Lloyd's algorithm \cite{lloyd1982} minimizes \eqref{eq:coverage_cost} with full observability of the environment. This centralized \textit{clairvoyant} algorithm is our primary benchmark for comparison.

The test-bed for the PAC system is a formulation of coverage control where the sensing and communication of each robot are restricted as illustrated in \fgref{fig:starling_hardware}. The environment is not known a priori to the multi-robot system. 
The LPAC policy we equipped on our robots is trained end-to-end to minimize \eqref{eq:coverage_cost} using imitation learning on state-action pairs generated by using the clairvoyant expert. While training is centralized, we deploy the policy in a distributed manner, with each robot running an identical configuration of our PAC system. 

We validate the system outdoors with our real-world fleet as well as in grounded simulations that are informed by our real-world experiments.
We generate $4$ random IDFs with $F$ Gaussian features for real-world experiments and an additional $10$ random IDFs that are in-distribution for the training policy; \tbref{tab:idfs}.
Mission areas are defined at two scales $\qty{32}{\meter} \times \qty{32}{\meter}$ (exclusively used for $N\leq 10$) and $\qty{64}{\meter} \times \qty{64}{\meter}$.
During training, LPAC operates at $\qty{5}{\meter\per\second}$ which we scale to \qty{1.25}{\meter\per\second} for deployments in smaller environments than seen during training.
The same velocity scaling of the policy's output is applied to each robot in both real-world and simulated experiments for consistency.
Each robot has a real-world $4 \times \SI{4}{\meter\squared}$ sensor area with which it explores the environment and we restrict message aggregation to a $\qty{16}{\meter}$ neighborhood.
\begin{table}[t]
    \centering
    \caption{Experiment IDFs.}
    \label{tab:idfs}
    \setlength{\tabcolsep}{4pt}
    \begin{tabular}{lccl}
      \toprule
      IDF & Gaussians $F$ & Mission Area & Usage \\
      \midrule
      Envs 1--3 & 10 & $\qty{32}{\meter} \times \qty{32}{\meter}$ & Real-world \\
      Env 4 & 20 & $\qty{64}{\meter} \times \qty{64}{\meter}$ & Real-world and Sim\\
      Envs 5--14 ($\times$10) & 32 & N/A & Simulation\\
      \bottomrule
    \end{tabular}
  \end{table}

\subsection{Global Execution from Local Communication}
\label{sec:c1}
\begin{figure}
    \includegraphics[width=\linewidth]{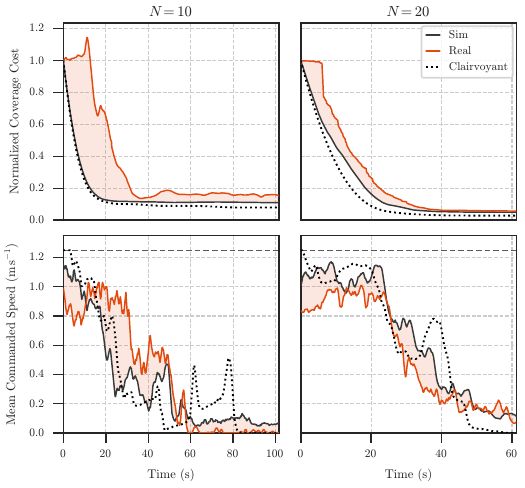}
    \caption{
        Sim-to-real comparison for identical initial conditions and configuration.
        \textbf{Top:} Normalized coverage cost comparison between simulated (black) and real-world LPAC controller (red).
        \textbf{Bottom:} Mean commanded velocity of the simulated and real system. The dashed line shows maximum command velocity.
    }
    \label{fig:simtoreal}
\end{figure}

The hallmark capability of our PAC system is its ability to approach the task performance of the clairvoyant expert it is trained on, while deployed as a decentralized system restricted to local sensing and communication.
We demonstrate that our PAC system embodies this capability at $N=F=10$ and $N=F=20$ in real-world outdoor experiments in environments of size $32 \times \qty{32}{\meter\squared}$ and $64 \times \qty{64}{\meter\squared}$, respectively.
The real-world trials are compared to a simulated variant of the LPAC system with the same information access but perfect communications and localization, and the clairvoyant expert which has complete task information.
We execute simulated trials with the same maximum velocity used in the real-world experiments, $\qty{1.25}{\meter\per\second}$.
These comparisons characterize the system's sim-to-real gap and the decentralized policy's capability to execute with global capability; \fgref{fig:simtoreal}.

The sim-to-real gap diminishes at larger scale, despite the system suffering from degraded communications as the network struggles to support the large fleet.
This illustrates the key capability of the PAC system to leverage an increased team size to obtain a complete task picture as a result of the diffusion of learned representations over the network.

\subsection{Scalability}
\label{sec:c2}
\begin{figure*}
  \includegraphics[trim={0.22cm, 0.30cm, 0.22cm, 0.22cm}, clip, width=\linewidth]{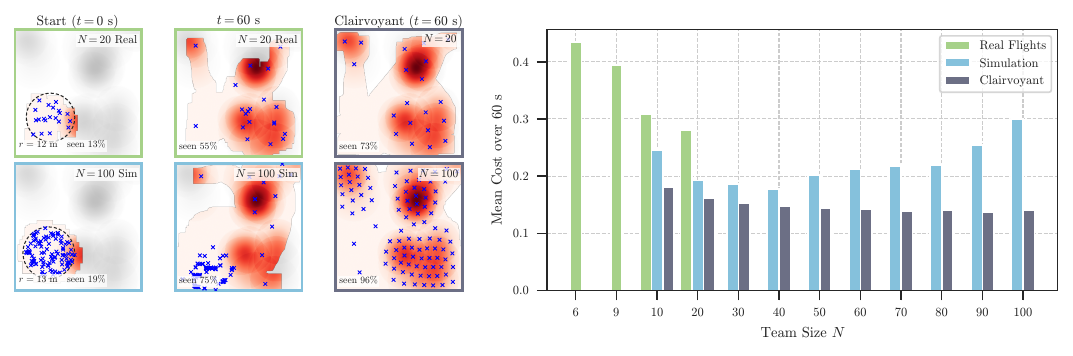}
    \caption{%
        Real-world LPAC experiments (green) are augmented with simulated LPAC trials (blue) at larger scales, with the clairvoyant for reference.
        \textbf{Left:} Team progression over $\qty{60}{\second}$ of real-world and simulated trials at $N=20$ and $N=100$ respectively.
        The blue $\times$ markers denote the positions of the robots in the environment, while the IDF is illustrated in red.
        The grayscale mass represents areas of the IDF that have not yet been explored by any robot. 
        \textbf{Right:} Cost across $N=6$ to $100$, highlighting the benefits and limitations of scalability for the policy.
        Simulated trials share real-world initial conditions where possible, and otherwise sample the real-world $N=20$ launch disk. \label{fig:scale-sim}}
\end{figure*}

Multi-robot systems benefit from the parallelism and redundancy inherent to homogeneous teams.
Scalable systems are required to leverage this advantage and we demonstrate that the proposed PAC framework is natively scalable both in the real-world and in simulation by conducting experiments on the same environment (Env \#4) with $F=20$ features, across various scales; shown in \fgref{fig:scale-sim}.
Real-world experiments supply the trials for $N = [6, 9, 10, 20]$ while simulated experiments enable projection of the system's capabilities and limitations when $N$ is scaled from $10$ to $100$ in intervals of $10$.
On a fixed size environment and with the same velocity limitations imposed in simulation, we find that the real-world experiments track the trend of improved coverage cost as $N$ increases.
The projection to scales beyond $N=20$ reveals that an optimum team size resides near $N=40$ before scale begins to be detrimental to LPAC's performance.
Notably, the clairvoyant baseline continues to improve but sees diminishing returns at scale, indicating that this is a limitation of the policy and not the local optimum for the task.
We remark that the initial conditions and the ratio of robots to features is far outside of the training distribution of LPAC~\cite{agarwal_2024}.

\subsection{Resilience to Team Reconfiguration}
\label{sec:c3}
Robot attrition is inevitable in fielded swarms and a system operating PAC loops must degrade gracefully when team members are lost to communication or vehicle failure.
We characterize this capability in simulation over a collection of 10 randomly generated environments with $F=32$, in which $N=32$ robots launch from a dense cluster that mimics the launch site of our real-world experiments.
In each environment, the team first runs to convergence, after which $m$ robots are instantly and permanently removed from the system and the surviving group continues the evaluation for \qty{60}{\second}.
We experiment with two orders of removing robots: interior first, which starts by removing robots near the center of mass of the team, and random, which captures arbitrary failures.

\begin{figure}
    \centering
    \includegraphics[width=\linewidth]{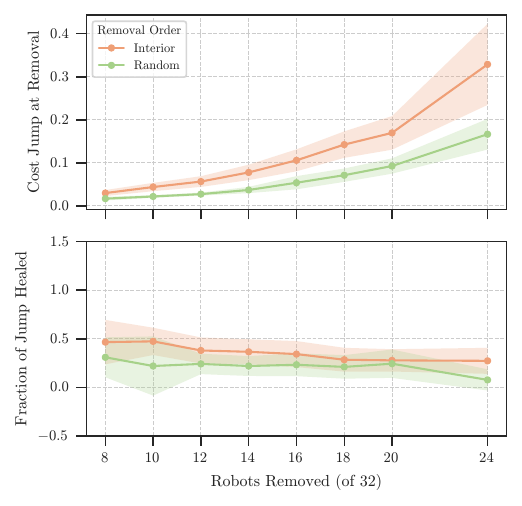}
    \caption{%
        Self-healing after convergence, aggregated over 10 randomly generated environments ($N=F=32$); darker lines are means, bands are $95\%$ CI.
        \textbf{Top:} The instantaneous cost jump when $m$ robots are permanently removed from a converged swarm.
        \textbf{Bottom:} The fraction of the jump recovered by the survivors declines for larger removals.}
    \label{fig:selfheal_removal}
\end{figure}

Removal is swept from $m = 8$ to $m = 24$ and we record the instantaneous jump in cost at removal, $\Delta\calJ = \calJ(t_r) - \calJ(t_r-1)$, where $t_r$ is the time at removal, as well as the fraction of the cost healed by the remaining members of the team; \fgref{fig:selfheal_removal}.
We find that as $m$ increases, the cost jump increases superlinearly in both removal cases, with structured removal near the center being more damaging than arbitrary removal.
Healing response, which is measured as $h = (\calJ(t_r) - \calJ(t_f))/\Delta\calJ$ where $t_f$ is the final timestep (\qty{60}{\second} after convergence), is strongest for moderate removals.
For $m=10$ on interior removal, the survivors are able to heal a mean of $46\%$ of the jump by expanding into the gaps left by the missing robots.
As $m$ increases, the recovery dampens, collapsing below $10\%$ for random removal at $m=24$.

\begin{figure}
    \centering
    \includegraphics[width=\linewidth]{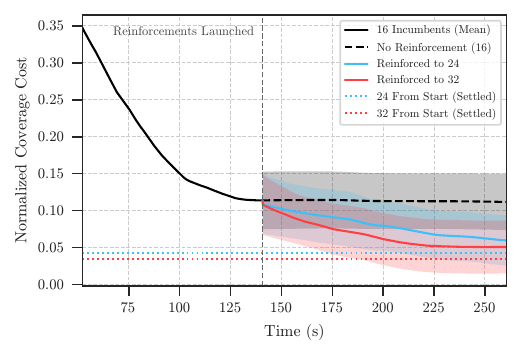}
    \caption{%
        Reinforcement of a converged swarm, aggregated over 10 randomly generated environments with $F=32$ features.
        The mean coverage cost of the $N=16$ incumbent team is shown up to the reinforcement time (dashed vertical line).
        Thereafter, arrivals launch from the original launch site and the branch traces are means with 95\% CI.
        Growing the team to $N=24$ or $N=32$ halves the settled cost, approaching the corresponding from-start settled costs (dotted lines).}
    \label{fig:reinforcement}
\end{figure}

Reinforcement is the dual of attrition and we study how the PAC system reacts when robots are added to an already converged swarm, substantially changing the team's configuration.
On the same group of environments as the removal experiments, we launch teams of $N=16$ incumbents and run to convergence.
Reinforcements of $M = [8, 16]$ are launched thereafter from the same launch sites and inherit no prior map.
The arrival of reinforcements enables the larger teams to reduce coverage cost even further, with the reinforcements integrating quickly and seamlessly into the existing team, shown in \fgref{fig:reinforcement}.
On average, the reinforced team approaches the coverage cost of the same team size launched at the same time.
Crucial to real-world deployments, the swarm absorbs the doubling of its size without any degradation in performance.

\subsection{Severely Degraded Communications}
\label{sec:c4}
When communication becomes severely degraded, robots must continue to make progress towards their objectives.
We evaluate this capability in simulation to instigate controlled communications failure. The initial configuration follows that of our real-world $20$ robot experiment.

\begin{figure}
    \centering
    \includegraphics[width=\linewidth]{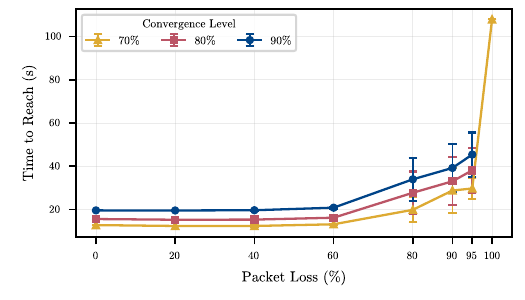}
    \caption{%
    Packet loss (PL) is swept from $0\%$ to $100\%$.
    We measure the time it takes for PAC to converge on a fraction of the coverage cost of PAC under no adverse PL (\texttt{no-PL}).
    Three convergence levels are measured: $70\%$ (yellow), $80\%$ (red) and $90\%$ (blue) with respect to the coverage cost of the \texttt{no-PL} scenario.
    The traces show the mean and $95\%$ CI over 10 trials where the solution is non-deterministic.
    At $100\%$ PL, the system only converges to $70\%$ of the \texttt{no-PL} coverage cost after $\qty{107}{\second}$.\label{fig:comms_dropout}}
\end{figure}

Modeled as an $N \times N$ Bernoulli matrix, we sweep the probability of a link dropout $p$ to model degraded communications under increasing packet loss (PL).
The active communication links are drawn and then combined with the spatial communication constraints to obtain a final communication mask for each timestep; \fgref{fig:comms_dropout}.
We measure the performance of the PAC system under degraded communications by comparing the time it takes to converge to $70, 80$ and $90\%$ of the unfettered, deterministic trial at $0\%$ PL (\texttt{no-PL}). 
At $N=20$ with a dense starting configuration, the redundant communication links are exploited by the GNN to keep the fleet on-task, even under high PL. %

\subsection{Repeatability and Transferability}
\label{sec:c5}
\begin{figure}
    \includegraphics[width=\linewidth]{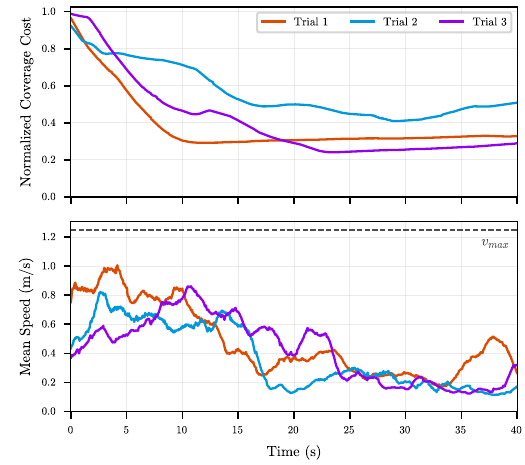}
    \caption{Repeatability is evaluated on Env \#1 (\tbref{tab:idfs}).
        \textbf{Top:} the normalized coverage cost of the three trials over $\qty{40}{\second}$; the trajectories vary and Trial \#2 struggles to converge.
        \textbf{Bottom:} The average speed across robots captured over time for each trial.
        The initial spike is followed by a gradual decay as robots converge on their final positions.}
    \label{fig:repeatable}
\end{figure}

\begin{figure}
    \centering
    \includegraphics[width=\linewidth]{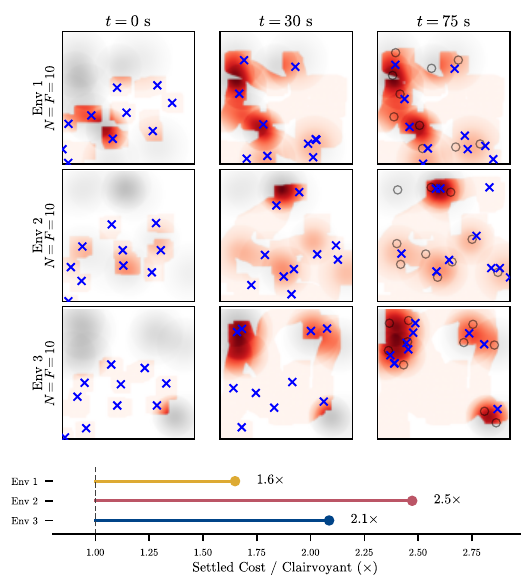}
    \caption{
        Performance of the LPAC system in the real world across three different environments in a $32 \times \SI{32}{\meter\squared}$ area.
        Each trial is executed with $N = 10$ robots and $F = 10$ Gaussian features.
        \textbf{Top:} Convergence occurs within \SI{75}{\second} and exact rates are dependent on environment and initial robot positions.
        The black circles represent the final positions of the clairvoyant run from the same initial conditions at the final timestep.
        \textbf{Bottom:} the final cost compared to the clairvoyant centralized algorithm.
    }
    \label{fig:transferability}
\end{figure}

To demonstrate transferability to novel environments, we first establish the repeatability of our PAC system, driven by a deterministic control policy, to remain consistent under the noisy conditions of real-world localization.
We run three trials on Env \#1 with the same launch positions for each robot and measure the coverage cost and mean speed over a $\qty{40}{\second}$ window; \fgref{fig:repeatable}.
Trials are time aligned by the mean fleet speed crossing a $\qty{0.3}{\meter\per\second}$ threshold once takeoff altitude is reached (later experiments use an operator signaling mechanism for experiment start).
Trials 1 and 3 obtain a similar final cost of $0.328$ and $0.291$, respectively, while Trial 2 has worse performance at a final cost of $0.509$.
Under the experimental conditions, the GPS localization error was typically between $1$ and $2$ meters resulting in significant noise.
This error is magnified when the robots are near the ground and GNSS suffers from signal blockage and multi-path interference.
Transferability of LPAC policies was established empirically in simulation~\cite{agarwal_2024}.
Here, we verify it in the real-world by deploying our system on unseen density functions (\tbref{tab:idfs}).
Maintaining consistent scale and initial conditions, we execute trials on Envs \#1 through \#3 with $N=10$ robots in succession.
Snapshots from these experiments are shown in \fgref{fig:transferability}.
In all three environments, the LPAC policy achieves coverage of the importance field with varying success compared to the clairvoyant.
We remark that the spread in final cost across unseen environments is comparable to the trial-to-trial spread shown in \fgref{fig:repeatable}, indicating that transferring the policy to novel IDFs has little cost beyond variances due to real-world noise.

\section{Conclusion}
\label{sc:conclusion}
We introduced a framework for asynchronous Perception-Action-Communication with Graph Neural Networks.
By acting as the bridge between high-level perception and low-level control, the GNN synthesizes collaborative messages that enable swarms to achieve global objectives through purely local interactions. Our asynchronous architecture decouples these subtasks, allowing GNN-based inference and message passing to operate at optimal frequencies without bottlenecking perception or action.

We validated this framework on the coverage control problem, demonstrating scalability across 100-agent simulations and physical deployment of 20 robots. Engineered atop ROS2 and the Zenoh network protocol for efficient embedding exchange, our system exhibited a remarkably narrow sim-to-real gap. The learned policies proved resilient to dynamic team sizes, varied initial conditions, and realistic network latency and packet loss.

Future work entails validating the system with even larger scale teams and fully decentralized communications. While physical deployments highlighted bottlenecks in network saturation and hardware platform limitations, they chart a clear path for future improvements such as transitioning to ad-hoc mesh networks and platforms designed for field and outdoor use. Because our asynchronous PAC framework is task-agnostic, future work also includes training and deploying GNN policies across other multi-robot challenges, including dynamic flocking and target tracking.

\bibliographystyle{IEEEtran}
\bibliography{IEEEabrv,LaTeX_macros_bib/ieeeabrv_plus,LaTeX_macros_bib/references}

\begin{thebibliography}{10}
\providecommand{\url}[1]{#1}
\csname url@samestyle\endcsname
\providecommand{\newblock}{\relax}
\providecommand{\bibinfo}[2]{#2}
\providecommand{\BIBentrySTDinterwordspacing}{\spaceskip=0pt\relax}
\providecommand{\BIBentryALTinterwordstretchfactor}{4}
\providecommand{\BIBentryALTinterwordspacing}{\spaceskip=\fontdimen2\font plus
\BIBentryALTinterwordstretchfactor\fontdimen3\font minus
  \fontdimen4\font\relax}
\providecommand{\BIBforeignlanguage}[2]{{%
\expandafter\ifx\csname l@#1\endcsname\relax
\typeout{** WARNING: IEEEtran.bst: No hyphenation pattern has been}%
\typeout{** loaded for the language `#1'. Using the pattern for}%
\typeout{** the default language instead.}%
\else
\language=\csname l@#1\endcsname
\fi
#2}}
\providecommand{\BIBdecl}{\relax}
\BIBdecl
\renewcommand{\BIBentryALTinterwordstretchfactor}{4}

\bibitem{agarwal_2024}
S.~Agarwal, R.~Muthukrishnan, W.~Gosrich, V.~Kumar, and A.~Ribeiro, ``{LPAC}:
  Learnable perception-action-communication loops with applications to coverage
  control,'' \emph{{IEEE} Trans. Robot.}, vol.~41, pp. 5986--6005, 2025.

\bibitem{hu2022}
T.-K. Hu \emph{et~al.}, ``Scalable perception-action-communication loops with
  convolutional and graph neural networks,'' \emph{{IEEE} Trans. Signal Inf.
  Process. Netw.}, vol.~8, pp. 12--24, 2022.

\bibitem{KipfW17}
T.~N. Kipf and M.~Welling,
  ``\href{https://openreview.net/forum?id=SJU4ayYgl}{Semi-supervised
  classification with graph convolutional networks},'' in \emph{Proc. Int.
  Conf. Learn. Represent.}, 2017.

\bibitem{bruna2014spectral}
J.~Bruna, W.~Zaremba, A.~Szlam, and Y.~LeCun, ``Spectral networks and locally
  connected networks on graphs,'' in \emph{Proc. Int. Conf. Learn. Represent.},
  2014.

\bibitem{deferrard2016convolutionalnnongraphs}
M.~Defferrard, X.~Bresson, and P.~Vandergheynst, ``Convolutional neural
  networks on graphs with fast localized spectral filtering,'' in \emph{Proc.
  Int. Conf. Neural Inf. Process. Syst.}, 2016, pp. 3837--3845.

\bibitem{TolstayaPMLR20}
E.~Tolstaya, F.~Gama, J.~Paulos, G.~Pappas, V.~Kumar, and A.~Ribeiro,
  ``Learning decentralized controllers for robot swarms with graph neural
  networks,'' in \emph{Proc. Conf. Robot Learn.}, 2020, pp. 671--682.

\bibitem{GamaDecentralized2022}
F.~Gama, Q.~Li, E.~Tolstaya, A.~Prorok, and A.~Ribeiro, ``Synthesizing
  decentralized controllers with graph neural networks and imitation
  learning,'' \emph{{IEEE} Trans. Signal Process.}, vol.~70, pp. 1932--1946,
  2022.

\bibitem{chen2024learning}
S.~Chen, Y.~Sun, P.~Li, L.~Zhou, and C.-T. Lu, ``Learning decentralized
  flocking controllers with spatio-temporal graph neural network,'' in
  \emph{Proc. {IEEE} Int. Conf. Robot. Autom.}, 2024, pp. 2596--2602.

\bibitem{Gosrich22}
W.~Gosrich \emph{et~al.},
  ``\href{https://doi.org/10.1109/ICRA46639.2022.9811854}{Coverage control in
  multi-robot systems via graph neural networks},'' in \emph{Proc. {IEEE} Int.
  Conf. Robot. Autom.}, 2022, pp. 8787--8793.

\bibitem{zhou2022}
L.~Zhou \emph{et~al.}, ``Graph neural networks for decentralized multi-robot
  target tracking,'' in \emph{Proc. {IEEE} Int. Symp. Saf., Secur., Rescue
  Robot.}, 2022, pp. 195--202.

\bibitem{rangesh2021trackmpnn}
A.~Rangesh, P.~Maheshwari, M.~Gebre, S.~Mhatre, V.~Ramezani, and M.~M. Trivedi,
  ``{TrackMPNN}: A message passing graph neural architecture for multi-object
  tracking,'' \emph{arXiv preprint arXiv:2101.04206}, 2021.

\bibitem{RuizGR21}
L.~Ruiz, F.~Gama, and A.~Ribeiro, ``Graph neural networks: Architectures,
  stability, and transferability,'' \emph{Proc. {IEEE}}, vol. 109, no.~5, pp.
  660--682, 2021.

\bibitem{BlumenkampGNN2022}
J.~Blumenkamp, S.~Morad, J.~Gielis, Q.~Li, and A.~Prorok, ``A framework for
  real-world multi-robot systems running decentralized {GNN}-based policies,''
  in \emph{Proc. {IEEE} Int. Conf. Robot. Autom.}, 2022, pp. 8772--8778.

\bibitem{Tzes_icra_2023}
M.~Tzes, N.~Bousias, E.~Chatzipantazis, and G.~J. Pappas, ``Graph neural
  networks for multi-robot active information acquisition,'' in \emph{Proc.
  {IEEE} Int. Conf. Robot. Autom.}, 2023, pp. 3497--3503.

\bibitem{ros2}
S.~Macenski, T.~Foote, B.~Gerkey, C.~Lalancette, and W.~Woodall, ``{Robot
  Operating System} 2: Design, architecture, and uses in the wild,'' \emph{Sci.
  Robot.}, vol.~7, no.~66, May 2022, {Art. no.} eabm6074.

\bibitem{Cortes05}
J.~Cort{\'e}s, S.~Mart{\'\i}nez, and F.~Bullo,
  ``\href{https://doi.org/10.1051/cocv:2005024}{Spatially-distributed coverage
  optimization and control with limited-range interactions},'' \emph{{ESAIM}
  Control Optim. Calc. Var.}, vol.~11, no.~4, pp. 691--719, 2005.

\bibitem{zenoh2023}
A.~Corsaro \emph{et~al.}, ``Zenoh: Unifying communication, storage and
  computation from the cloud to the microcontroller,'' in \emph{Proc. Euromicro
  Conf. Digit. Syst. Des.}, 2023, pp. 422--428.

\bibitem{Karney_2012}
C.~F.~F. Karney, ``Algorithms for geodesics,'' \emph{J. Geod.}, vol.~87, no.~1,
  pp. 43--55, Jan. 2013.

\bibitem{eProsima_Micro_XRCE_DDS}
{eProsima}, ``{eProsima Micro XRCE-DDS},''
  \url{https://micro-xrce-dds.docs.eprosima.com/}, 2025, documentation v3.0.1.
  Accessed: Sep. 23, 2026.

\bibitem{ModalAI_VOXL2_product}
{ModalAI, Inc.}, ``{VOXL} 2,'' \url{https://www.modalai.com/products/voxl-2},
  product page. Accessed: Sep. 23, 2026.

\bibitem{Alfa_AWUS036EACS_product}
{ALFA Network Inc.}, ``{AWUS036EACS}: 802.11ac {WiFi} + {BT} {USB} adapter,''
  \url{https://www.alfa.com.tw/products/awus036eacs_1}, product page. Accessed:
  Sep. 23, 2026.

\bibitem{Du99}
Q.~Du, V.~Faber, and M.~Gunzburger,
  ``\href{https://doi.org/10.1137/S0036144599352836}{Centroidal Voronoi
  tessellations: Applications and algorithms},'' \emph{{SIAM} Rev.}, vol.~41,
  no.~4, pp. 637--676, 1999.

\bibitem{lloyd1982}
S.~P. Lloyd, ``Least squares quantization in {PCM},'' \emph{{IEEE} Trans. Inf.
  Theory}, vol.~28, no.~2, pp. 129--137, 1982.

\end{thebibliography}
\end{document}